\documentclass{article}

\usepackage{arxiv}

\usepackage[utf8]{inputenc} 
\usepackage[T1]{fontenc}    
\usepackage{hyperref}       
\usepackage{url}            
\usepackage{booktabs}       
\usepackage{amsfonts}       
\usepackage{nicefrac}       
\usepackage{microtype}      
\usepackage{lipsum}
\usepackage{multirow}
\usepackage{makecell}
\usepackage{tabularx}
\usepackage{graphicx}
\usepackage{caption}
\usepackage{enumitem}
\usepackage{textcomp}
\usepackage[page]{appendix}

\graphicspath{ {include/} }

\title{Pre-training Polish Transformer-based Language Models at Scale}

\author{
 Sławomir Dadas \\
  National Information Processing Institute\\
  Warsaw, Poland\\
  \texttt{sdadas@opi.org.pl} \\
   \And
 Michał Perełkiewicz \\
  National Information Processing Institute\\
  Warsaw, Poland\\
  \texttt{mperelkiewicz@opi.org.pl} \\
  \And
 Rafał Poświata \\
  National Information Processing Institute\\
  Warsaw, Poland\\
  \texttt{rposwiata@opi.org.pl} \\
}

\begin{document}
\maketitle

\begin{abstract}
Transformer-based language models are now widely used in Natural Language Processing (NLP). This statement is especially true for English language, in which many pre-trained models utilizing transformer-based architecture have been published in recent years. This has driven forward the state of the art for a variety of standard NLP tasks such as classification, regression, and sequence labeling, as well as text-to-text tasks, such as machine translation, question answering, or summarization. The situation have been different for low-resource languages, such as Polish, however. Although some transformer-based language models for Polish are available, none of them have come close to the scale, in terms of corpus size and the number of parameters, of the largest English-language models. In this study, we present two language models for Polish based on the popular BERT architecture. The larger model was trained on a dataset consisting of over 1 billion polish sentences, or 135GB of raw text. We describe our methodology for collecting the data, preparing the corpus, and pre-training the model. We then evaluate our models on thirteen Polish linguistic tasks, and demonstrate improvements over previous approaches in eleven of them.
\end{abstract}

\section{Introduction}
Unsupervised pre-training for Natural Language Processing (NLP) has gained popularity in recent years. The goal of this approach is to train a model on a large corpus of unlabeled text, and then use the representations the model generates as an input for downstream linguistic tasks. The initial popularization of these methods was related to the successful applications of pre-trained word vectors (embeddings), the most notable of which include Word2Vec \citep{mikolov2013distributed}, GloVe \citep{pennington2014glove}, and FastText \citep{bojanowski2017enriching}. These representations have contributed greatly to the development of NLP. However, one of the main drawbacks of such tools was that the static word vectors did not encode contextual information. The problem was addressed in later studies by proposing context-dependent representations of words based on pre-trained neural language models. For this purpose, several language model architectures which utilize bidirectional long short-term memory (LSTM) layers have been introduced. The popular models such as ELMo \citep{peters2018deep}, ULMFiT \citep{howard2018universal}, and Flair \citep{akbik2018contextual}, have led to significant improvements in a wide variety of linguistic tasks. Shortly after, \citet{devlin-etal-2019-bert} introduced BERT - a different type of language model based on transformer \citep{vaswani2017attention} architecture. Instead of predicting the next word in a sequence, BERT is trained to reconstruct the original sentence from one in which some tokens have been replaced by a special \emph{mask token}. Since the text representations generated by BERT have proved to be effective for NLP problems - even those which were previously considered challenging, such as question answering or common sense reasoning - more focus has been put on transformer-based language models. As a result, in the last two years we have seen a number of new methods based on that idea, with some modifications in the architecture or the training objectives. The approaches that have gained wide recognition include RoBERTa \citep{liu2019roberta}, Transformer-XL \citep{dai-etal-2019-transformer}, XLNet \citep{yang2019xlnet}, Albert \citep{lan2019albert}, and Reformer \citep{kitaev2019reformer}.

The vast majority of research on both transformer-based language models and transfer learning for NLP is targeted toward the English language. This progress does not translate easily to other languages. In order to benefit from recent advancements, language-specific research communities must adapt and replicate studies conducted in English to their native languages. Unfortunately, the cost of training state-of-the-art language models is growing rapidly \citep{peng_2019}, which makes not only individual scientists, but also some research institutions unable to reproduce experiments in their own languages. Therefore, we believe that it is particularly important to share the results of research - especially pre-trained models, datasets, and source code of the experiments - for the benefit of the whole scientific community. In this article, we describe our methodology for training two language models for Polish language based on BERT architecture. The smaller model follows the hyperparameters of an English-language BERT-base model, and the larger version follows the BERT-large model. To the best of our knowledge, the latter is the largest language model for Polish available to date, both in terms of the number of parameters (355M) and the size of the training corpus (135GB). We have released both pre-trained models publicly\footnote{\url{https://github.com/sdadas/polish-roberta}}. We evaluate our models on several linguistic tasks in Polish, including nine from the KLEJ benchmark \citep{rybak2020klej}, and four additional tasks. The evaluation covers a set of typical NLP problems, such as binary and multi-class classification, textual entailment, semantic relatedness, ranking, and Named Entity Recognition (NER).

\subsection{Language-specific and multilingual transformer-based models}
In this section we provide an overview of models based on the transformer architecture for languages other than English. Apart from English, the language on which NLP research is most focused currently is Chinese. This is reflected in the number of pre-trained models available \citep{xu2020cluecorpus2020,chinese-bert-wwm,devlin-etal-2019-bert,sun2019ernie}. Other languages for which we found publicly available pre-trained models included: Arabic \citep{antoun2020arabert}, Dutch \citep{de2019bertje,delobelle2020robbert}, Finnish \citep{virtanen2019multilingual}, French \citep{martin2019camembert,le2019flaubert}, German, Greek, Italian, Japanese, Korean, Malaysian, Polish, Portuguese \citep{souza2019portuguese}, Russian \citep{kuratov2019adaptation}, Spanish \citep{CaneteCFP2020}, Swedish, Turkish, and Vietnamese \citep{nguyen2020phobert}. Models covering a few languages of the same family are also available, such as SlavicBERT (Bulgarian, Czech, Polish, and Russian) \citep{arkhipov-etal-2019-tuning} and NordicBERT\footnote{\url{https://github.com/botxo/nordic_bert}} (Danish, Norwegian, Swedish, and Finnish). The topic of massive multilingual models covering tens, or in some cases more than a hundred languages, has attracted more attention in recent years. The original BERT model \citep{devlin-etal-2019-bert} was released along with a multilingual version covering 104 languages. XLM \citep{NIPS2019_8928} (fifteen, seventeen and 100 languages) and XLM-R \citep{conneau2019unsupervised} (100 languages) were released in 2019. Although it was possible to use these models for languages in which no monolingual models were available, language-specific pre-training usually leads to better performance. To date, two BERT-base models have been made available for Polish: HerBERT \citep{rybak2020klej} and Polbert\footnote{\url{https://github.com/kldarek/polbert}}, both of which utilize BERT-base architecture.

\subsection{Contributions} 
Our contributions are as follows: 
1) We trained two transformer-based language models for Polish, consistent with the BERT-base and BERT-large architectures. To the best of our knowledge, the second model is the largest language model trained for Polish to date, both in terms of the number of parameters and the size of the training corpus. 
2) We proposed a method for collecting and pre-processing the data from the Common Crawl database to obtain clean, high-quality text corpora. 
3) We conducted a comprehensive evaluation of our models on thirteen Polish linguistic tasks, comparing them to other available transformer-based models, as well as recent state-of-the-art approaches.
4) We made the source code of our experiments available to the public, along with the pre-trained models. 

\section{Language model pre-training}
\label{sec:methodology}
In this section, we describe our methodology for collecting and pre-processing the data used for training BERT-base language models. We then present the details of the training, explaining our procedure and the selection of hyperparameters used in both models.

\subsection{Training corpus}
Transformer-based models are known for their high capacity \citep{jawahar:hal-02131630,kovaleva-etal-2019-revealing}, which means that they can benefit from large quantities of text. An important step in the process of creating a language model, therefore, is to collect a sufficiently large text corpus. We have taken into account that the quality of the text used for training will also affect the final performance of the model. The easiest way to collect a large language-specific corpus is to extract it from Common Crawl - a public web archive containing petabytes of data crawled from web pages. The difficulty with this approach is that web-based data is often noisy and unrepresentative of typical language use, which could eventually have a negative impact on the quality of the model. In response to this, we have developed a procedure for filtering and cleaning the Common Crawl data to obtain a high-quality web corpus. The procedure is as follows:
\begin{enumerate}[labelwidth=0pt, labelindent=0pt]
\item We download full HTML pages (WARC files in Common Crawl), and use the resulting metadata to filter the documents written in Polish language.
\item We use \emph{Newspaper3k}\footnote{\url{https://newspaper.readthedocs.io/en/latest/}} - a tool which implements a number of heuristics for extracting the \emph{main content} of the page, discarding any other text such as headers, footers, advertisements, menus, or user comments.
\item We then remove all texts shorter than 100 characters. Additionally, we identify documents containing the words: ‘przeglądarka’, ‘ciasteczka’, ‘cookies’, or ‘javascript’. The presence of these words may indicate that the extracted content is a description of a cookie policy, or default content for browsers without JavaScript enabled. We discard all such texts if they are shorter than 500 characters.
\item In the next step, we use a simple statistical language model (KenLM \citep{heafield2011kenlm}), trained on a small Polish language corpus to assess the quality of each extracted document. For each text, we compute the perplexity value and discard all texts with perplexity higher than 1000.
\item Finally, we remove all duplicated texts.
\end{enumerate}

The full training corpus we collected is approximately 135GB in size, and is composed of two components: the web part and the base part. For the web part, which amounts to 115GB of the corpus, we downloaded three monthly dumps of Common Crawl data, from November 2019 to January 2020, and followed the pre-processing steps described above. The base part, which comprises the remaining 20GB, is composed of publicly available Polish text corpora: the Polish language version of Wikipedia (1.5GB), the Polish Parliamentary Corpus (5GB), and a number of smaller corpora from the CLARIN (\url{http://clarin-pl.eu}) and OPUS (\url{http://opus.nlpl.eu}) projects, as well as Polish books and articles.

\subsection{Training procedure}

The authors of the original BERT paper \citep{devlin-etal-2019-bert} proposed two versions of their transformer-based language model: BERT-large (more parameters and higher computational cost), and BERT-base (fewer parameters, more computationally efficient). To train the models for Polish language, we adapted the same architectures. Let $L$ denote the number of encoder blocks, $H$ denote the hidden size of the token representation, and $A$ denote the number of attention heads. Specifically, we used $L=12,H=768, A=12$ for the base model, and $L=24,H=1024, A=16$ for the large model. The large model was trained on the full 135GB text corpus, and the base model on only the 20GB base part. The training procedure we employed is similar to the one suggested in the RoBERTa pre-training approach \citep{liu2019roberta}. Originally, BERT utilized two training objectives - Masked Language Modeling (MLM), and Next Sentence Prediction (NSP). We trained our models with the MLM objective, since it has been shown that NSP fails to improve the performance of the pre-trained models on downstream tasks \cite{liu2019roberta}. We also used dynamic token masking, and trained the model with a larger batch size than the original BERT. The base model was trained with a batch size of 8000 sequences for 125 000 training steps: the large model was trained with a batch size of 30 000 sequences for 50 000 steps. The reason for using such a large batch size for the bigger model is to stabilize the training process. During our experiments, we observed significant variations in training loss for smaller batch sizes, indicating that the initial combination of learning rate and batch size had caused an exploding gradient problem. To address the issue, we increased the batch size until the loss stabilized.

Both models were pre-trained with the Adam optimizer using the following optimization hyperparameters: $\epsilon = 1\mathrm{e}{-6},\beta_{1} = 0.9,\beta_{2} = 0.98$. We utilized a learning rate scheduler with linear decay. The learning rate is first increased for a warm-up phase of 10 000 update steps to reach a peak of $7\mathrm{e}{-4}$, and then linearly decreased for the remainder of the training. We also mimicked the dropout approach of the original BERT model: a dropout of 0.1 is applied on all layers and attention weights. The maximum length of a sequence was set to 512 tokens. We do not combine sentences from the training corpus: each is treated as a separate training sample. To encode input sequences into tokens, we employed SentencePiece \citep{kudo-richardson-2018-sentencepiece} Byte Pair Encoding (BPE) algorithm, and set the maximum vocabulary size to 50 000 tokens.

\section{Evaluation}
\label{sec:evaluation}
In this section, we discuss the process and results of evaluating our language models on thirteen Polish downstream tasks. Nine of these tasks constitute the recently developed KLEJ benchmark \citep{rybak2020klej}; three of them have already been introduced in \citet{dadas-etal-2020-evaluation}; and the last, named entity recognition, was a part of the PolEval\footnote{\url{http://2018.poleval.pl/index.php/tasks}} evaluation challenge. First, we compare the performance of our models with other Polish and multilingual language models evaluated on the KLEJ benchmark. Next, we present detailed per-task results, comparing our models with the previous state-of-the-art solutions for each of the tasks.

\subsection{Task descriptions}
\textbf{NKJP} 
(The National Corpus of Polish (Narodowy Korpus Języka Polskiego)) \cite{przep-2012} is one of the largest text corpora of the Polish language, consisting of texts from Polish books, news articles, web content, and transcriptions of spoken conversations. A part of the corpus, known as the ‘one million subcorpus’, contains annotations of named entities from six categories: ‘persName’, ‘orgName’, ‘geogName’, ‘placeName’, ‘date’, and ‘time’. The authors of the KLEJ benchmark used this subset to create a named entity classification task \cite{rybak2020klej}. First, they filtered out all sentences containing entities of more than one type. Next, they randomly assigned sentences to train development and test sets according to the rule that each named entity mentioned appears in only one of these splits. They undersample the ‘persName’ class, and merge the ‘date’ and ‘time’ classes to increase class balance.  Finally, they selected sentences without any named entity, and assigned them the ‘noEntity’ label. The resulting dataset consisted of 20 000 sentences belonging to six classes. The task is to predict the presence and type of each named entity. Classification accuracy is also reported.

\textbf{8TAGS} 
is a corpus created by \citet{dadas-etal-2020-evaluation} for their study on the subject of sentence representations in Polish language. This dataset was created automatically by extracting sentences from headlines and short descriptions of articles posted on the Polish social network, wykop.pl. It contains approximately 50 000 sentences, all longer than thirty characters, from eight popular categories: film, history, food, medicine, automotive, work, sport, and technology. The task is to assign a sentence to one of these classes in which classification accuracy is the measure.

\textbf{CBD} 
(Cyberbullying Detection) \cite{ptaszynski2019results} is a binary classification task, the goal of which is to determine whether a Twitter message constitutes a case of cyberbullying or not. This was a sub-task of task 6 in the PolEval 2019 competition. The dataset prepared by the competition’s organizers contains 11 041 tweets, extracted from nineteen of the most popular Polish Twitter accounts in 2017. The F1-score was used to measure the performance of the models.

\textbf{DYK}
‘Did you know?’ (\textit{‘Czy wiesz?’}) \cite{MarPtaRadzPia:13} is a dataset used for the evaluation and development of Polish language question answering systems. It consists of 4721 question-answer pairs obtained from the \textit{Czy wiesz...} Polish Wikipedia project. The answer to each question was found in the linked Wikipedia article. \citet{rybak2020klej} used this dataset to devise a binary classification task, the goal of which is to predict whether the answer to the given question is correct or not \cite{rybak2020klej}. Positive responses were additionally marked within larger fragments of responded text. Negative samples were selected by the BPE token overlap between a question and a possible answer. The F1-score was also reported for this task.

\textbf{PSC} 
The Polish Summaries Corpus \cite{OGRODNICZUK14.1211} is a corpus of manually created summaries of Polish language news articles. The dataset contains both abstract free-word summaries and extraction-based summaries created by selecting text spans from the original documents. Based on PSC, \cite{rybak2020klej} formulated a text-similarity task \cite{rybak2020klej}. They generate positive pairs by matching each extractive summary with the two least similar abstractive ones in the same article. Negative pairs were obtained by finding the two most similar abstractive summaries for each extractive summary, but from different articles. To calculate the similarity between summaries, they used the BPE token overlap. The F1-score was used for evaluation.

\textbf{PolEmo2.0} 
 \cite{kocon-etal-2019-multi-level} is a corpus of consumer reviews obtained from 
four domains: medicine, hotels, products, and school. Each of the reviews is annotated with one of four labels: positive, negative, neutral, or ambiguous. In general, the task is to choose the correct label, although here two special versions of the task are distinguished: PolEmo2.0-IN and PolEmo2.0-OUT. In PolEmo2.0-IN, both the training and test sets come from the same domains, namely medicine and hotels. In PolEmo2.0-OUT, however, the test set comes from the product and school domains. In both cases, accuracy was used for evaluation.

\textbf{Allegro Reviews} 
(AR) \cite{rybak2020klej} is a sentiment analysis dataset of product reviews from the e-commerce marketplace, allegro.pl. Each review has a rating on a five-point scale, in which one is negative, and five is positive. The task is to predict the rating of a given review. The macro-average of the mean absolute error per class (wMAE) is applied for evaluation.

\textbf{CDSC} 
(The Compositional Distributional Semantics Corpus) \cite{wroblewska-krasnowska-kieras-2017-polish} is a corpus of 10 000 human-annotated sentence pairs for semantic relatedness and entailment, in which image captions from forty-six thematic groups were used as sentences. Two tasks are proposed based on this dataset.
The CDSC-R problem involves predicting the relatedness between a pair of sentences, on a scale of zero to five, in which zero indicates that the sentences are not related, and five indicates that they are highly related. In this task, the Spearman correlation is used as an evaluation measure. CDSC-E’s task is to classify whether the premise entails the hypothesis (entailment), negates the hypothesis (contradiction), or is unrelated (neutral). For this task, accuracy is reported.

\textbf{SICK} 
 \cite{dadas-etal-2020-evaluation} is a manually translated Polish language version of the English
Natural Language Inference (NLI) corpus, SICK (Sentences Involving Compositional Knowledge) \cite{marelli-etal-2014-sick}, and consists of 10 000 sentence pairs. As with the CDSC dataset, two tasks can also be distinguished here. SICK-R is the task of predicting the probability distribution of relatedness scores (ranging from 1 to 5) for the sentence pair, in which the Spearman correlation is used for evaluation. SICK-E is a multiclass classification problem in which the relationship between two sentences is classified as entailment, contradiction, or neutral. Accuracy is used once again to measure performance.

\textbf{PolEval-NER 2018} 
 \cite{poleval2018-ner} was task 2 in the PolEval 2018 competition, the goal of which was to detect and assign the correct category and subcategory (if applicable) to a found named entity. In this study the task was simplified, as only the main categories had to be found. The effectiveness of the models is verified by the F1-score measure. This task was prepared on the basis of the NKJP dataset previously presented.

\subsection{Task-specific fine-tuning}

To evaluate our language models on downstream tasks, we fine-tuned them separately for each task. In our experiments, we encounter three types of problem: classification, regression, and Named Entity Recognition (NER). In classification tasks, the model is expected to predict a label from a set of two or more classes. Regression concerns the prediction of a continuous numerical value. NER is a special case of sequence tagging, i.e. predicting a label for each element in a sequence. The dataset for each problem consists of training and test parts, and in most cases also includes a validation part. The general fine-tuning procedure is as follows: we train our model on the training part of the dataset for a specific number of epochs. If the validation set is available, we compute the validation loss after each epoch, and select the model checkpoint with the best validation loss. For datasets without a validation set, we select the last epoch checkpoint. Then, we perform an evaluation on the test set using the selected checkpoint.

In the case of classification and regression tasks, we attach an additional fully-connected layer to the output of the \emph{[CLS]} token, which always remains in the first position of a sequence. For classification, the number of outputs for this layer is equal to the number of classes, and the softmax activation function is used. For regression, it is a linear layer with a single output. The models are fine-tuned with the Adam optimizer using the following hyperparameters: $\epsilon = 1\mathrm{e}{-6},\beta_{1} = 0.9,\beta_{2} = 0.98$. A learning rate scheduler with polynomial decay is utilized. The first 6\% of the training steps are reserved for the warm-up phase, in which the learning rate is gradually increased to reach a peak of $1\mathrm{e}{-5}$. By default, we train for ten epochs with a batch size of sixteen sequences. The specific fine-tuning steps and exceptions to the procedure are discussed below:
\begin{itemize}[leftmargin=*, labelwidth=0pt, labelindent=0pt]
\item \textbf{Classification on imbalanced datasets} – Some of the binary classification datasets considered in the evaluation, such as CBD, DYK, and PSC, are imbalanced, which means that they contain significantly fewer samples of the first class than of the second class. To counter this imbalance, we utilize a simple resampling technique: samples for the minority class in the training set are duplicated, and some samples for the majority class are randomly discarded. We set the resampling factor to 3 for the minority class, and 1 (DYK, PSC) or 0.75 (CBD) respectively for the majority class. Additionally, we increase the batch size for those tasks to thirty-two. 
\item \textbf{Regression} - In many cases, a regression task is restricted to a specific range of values for which the prediction is valid. For example, Allegro Reviews contains user reviews with ratings between one and five stars. For fine-tuning, we scale all the outputs of regression models to be within the range of $[0,1]$, and then rescale them to their original range during evaluation. Before rescaling, any negative prediction is set to 0, and any prediction greater that 1 is limited to 1. 
\item \textbf{Named entity recognition} - Since sequence tagging, in which the model is expected to generate per-token predictions, is different from simple classification or regression tasks, we decided to adapt an existing named entity recognition approach for fine-tuning using our language models. For this purpose, we employed a method from \citet{shibuya2019nested}, who proposed a transformer-based named entity recognition model with a Conditional Random Fields (CRF) inference layer, and multiple Viterbi-decoding steps to handle nested entities. In our experiments, we used the same hyperparameters as the authors.
\end{itemize}

\begin{table}
  \centering
  \caption{Results on the KLEJ benchmark.}
  \aboverulesep=0ex
  \belowrulesep=0ex
  \begin{tabular}{l|c|ccccccccc}
    \toprule
    \textbf{Model} & \textbf{Average} & NKJP & CDSC-E & CDSC-R & CBD & PE2-I & PE2-O & DYK & \makecell[{{p{0.7cm}}}]{\centering PSC} & \makecell[{{p{0.7cm}}}]{\centering AR} \\
    \hline
    \multicolumn{11}{l}{\textbf{Base models}}\\
    \hline
    \makecell[l]{mBERT} & 79.5 & 91.4 & 93.8 & 92.9 & 40.0 & 85.0 & 66.6 & 64.2 & 97.9 & 83.3 \\
    \makecell[l]{SlavicBERT} & 79.8	& 93.3 & 93.7 & 93.3 & 43.1 & 87.1 & 67.6 & 57.4 & 98.3 & 84.3 \\
    \makecell[l]{XLM-100} & 79.9 & 91.6 & 93.7 & 91.8 & 42.5 & 85.6 & 69.8 & 63.0 & 96.8 & 84.2 \\
    \makecell[l]{XLM-17} & 80.2 & 91.9 & 93.7 & 92.0 & 44.8 & 86.3 & 70.6 & 61.8 & 96.3 & 84.5 \\
    \makecell[l]{HerBERT} & 80.5 & 92.7 & 92.5 & 91.9 & 50.3 & 89.2 & 76.3 & 52.1 & 95.3 & 84.5 \\
    \makecell[l]{XLM-R base} & 81.5	& 92.1 & 94.1 & 93.3 & 51.0 & 89.5 & 74.7 & 55.8 & 98.2 & 85.2 \\
    \makecell[l]{Polbert} & 81.7 & 93.6 & 93.4 & 93.8 & 52.7 & 87.4 & 71.1 & 59.1 & 98.6 & 85.2 \\
    \textbf{Our model} & \textbf{85.3} & \textbf{93.9} & \textbf{94.2} & \textbf{94.0} & \textbf{66.7} & \textbf{90.6} & \textbf{76.3} & \textbf{65.9} & \textbf{98.8} & \textbf{87.8} \\
    \hline
    \multicolumn{11}{l}{\textbf{Large models}}\\
    \hline
    \makecell[l]{XLM-R large} & 87.5 & 94.1 & \textbf{94.4} & 94.7 & 70.6 & 92.4 & 81.0 & 72.8 & \textbf{98.9} & 88.4 \\
    \textbf{Our model} & \textbf{87.8} & \textbf{94.5} & 93.3 & \textbf{94.9} & \textbf{71.1} & \textbf{92.8} & \textbf{82.4} & \textbf{73.4} & 98.8 & \textbf{88.8} \\
    \bottomrule
  \end{tabular}
  \label{tab:eval_klej}
\end{table}

\subsection{Results and discussion}
In this section, we demonstrate the results of evaluating our language models on downstream tasks. We repeated the fine-tuning of the models for each task five times. The scores reported are the median values of those five runs. Table \ref{tab:eval_klej} demonstrates the evaluation results on the KLEJ benchmark, in comparison with other available Polish and multilingual transformer-based models. The results of other approaches are taken from the KLEJ leaderboard. We split the table into two sections, comparing the BERT-base and BERT-large architectures separately. We can observe that there is a wider selection of base models, and most of them are multilingual, such as the original multilingual BERT (mBERT) \citep{devlin-etal-2019-bert}, SlavicBERT \citep{arkhipov-etal-2019-tuning}, XLM \citep{NIPS2019_8928}, and XLM-R \citep{conneau2019unsupervised}. The only models pre-trained specifically for Polish language are HerBERT \citep{rybak2020klej} and Polbert. Among the base models, our approach outperforms others by a significant margin. In the case of large models, only the XLM-RoBERTa (XLM-R) pre-trained model has been available until now. XLM-RoBERTa is a recently published multilingual transformer trained on 2.5TB of data in 100 languages. It has been shown to be highly competitive against monolingual models. A direct comparison with our Polish language model demonstrates a consistent advantage of our model - it has achieved better results in seven of the nine tasks included in the KLEJ benchmark.

\begin{table}
  \centering
  \caption{Detailed results for Polish language downstream tasks. In some cases, we used the datasets and task definitions from the KLEJ benchmark, which are different from the original tasks they were based on (they have been reformulated or otherwise modified by the benchmark authors). We denote such tasks with (KLEJ) to emphasize that the evaluation was performed on the KLEJ version of the data. The abbreviated task types are: C - classification, R - regression, and ST - sequence tagging.}
  \aboverulesep=0ex
  \belowrulesep=0ex
  \begin{tabular}{lr|c|lc|c|c}
    \toprule
    \textbf{Task} & \makecell[{{p{0.7cm}}}]{~} & \textbf{Metric} & \textbf{Previous SOTA} & \makecell[{{p{0.9cm}}}]{~} & \makecell[c]{\textbf{Base}\\ \textbf{model}} & \makecell[c]{\textbf{Large}\\ \textbf{model}}\\
    \hline
    \multicolumn{5}{l}{\textbf{Multi-class classification}}\\
    \hline
    \makecell[l]{NKJP (KLEJ)} & \textbf{C} & Accuracy & XLM-R large \citep{conneau2019unsupervised} & 94.1 & 93.9 (\textminus0.2) & \textbf{94.5} (+0.4) \\
    \makecell[l]{8TAGS} & \textbf{C} & Accuracy & ELMo \citep{dadas-etal-2020-evaluation} & 71.4 & 77.2 (+5.8) & \textbf{80.8} (+9.4) \\
    \hline
    \multicolumn{5}{l}{\textbf{Binary classification}}\\
    \hline
    \makecell[l]{CBD} & \textbf{C} & F1-score & XLM-R large \citep{conneau2019unsupervised} & 70.6 & 66.7 (\textminus2.9) & \textbf{71.1} (+0.5)  \\
    \makecell[l]{DYK (KLEJ)} & \textbf{C} & F1-score & XLM-R large \citep{conneau2019unsupervised} & 72.8 & 65.9 (\textminus6.9) & \textbf{73.4} (+0.6) \\
    \makecell[l]{PSC (KLEJ)} & \textbf{C} & F1-score & XLM-R large \citep{conneau2019unsupervised} & \textbf{98.9} & 98.8 (\textminus0.1) & 98.8 (\textminus0.1) \\
    \hline
    \multicolumn{5}{l}{\textbf{Sentiment analysis}}\\
    \hline
    \makecell[l]{PolEmo2.0-IN} & \textbf{C} & Accuracy & XLM-R large \citep{conneau2019unsupervised} & 92.4 & 90.6 (\textminus1.8) & \textbf{92.8} (+0.4) \\
    \makecell[l]{PolEmo2.0-OUT} & \textbf{C} & Accuracy & XLM-R large \citep{conneau2019unsupervised} & 81.0 & 76.3 (\textminus4.7) & \textbf{82.4} (+1.4) \\
    \makecell[l]{Allegro Reviews} & \textbf{R} & 1-wMAE & XLM-R large \citep{conneau2019unsupervised} & 88.4 & 87.8 (\textminus1.0) & \textbf{88.8} (+0.4) \\
    \hline
    \multicolumn{5}{l}{\textbf{Textual entailment}}\\
    \hline
    \makecell[l]{CDSC-E} & \textbf{C} & Accuracy & XLM-R large \citep{conneau2019unsupervised} & \textbf{94.4} & 94.2 (\textminus0.2) & 93.3 (\textminus1.1)  \\
    \makecell[l]{SICK-E} & \textbf{C} & Accuracy & LASER \citep{dadas-etal-2020-evaluation} & 82.2 & 86.1 (+3.9) & \textbf{87.7} (+5.5) \\
    \hline
    \multicolumn{5}{l}{\textbf{Semantic relatedness}}\\
    \hline
    \makecell[l]{CDSC-R} & \textbf{R} & Spearman & XLM-R large \citep{conneau2019unsupervised} & 94.7 & 94.0 (\textminus0.7) & \textbf{94.9} (+0.2) \\
    \makecell[l]{SICK-R} & \textbf{R} & Spearman & USE \citep{dadas-etal-2020-evaluation} & 75.8 & 82.3 (+6.5) & \textbf{85.6} (+9.8) \\
    \hline
    \multicolumn{5}{l}{\textbf{Named entity recognition}}\\
    \hline
    \makecell[l]{Poleval-NER 2018} & \textbf{ST} & F1-score & \citet{dadas2019combining} & 86.2 & 87.9 (+1.7) & \textbf{90.0} (+3.8) \\
    \bottomrule
  \end{tabular}
  \label{tab:eval_tasks}
\end{table}

Table \ref{tab:eval_tasks} shows a more detailed breakdown of the evaluation results, and includes all the tasks from the KLEJ benchmark, and four additional tasks: SICK-R, SICK-R, 8TAGS, and PolEval-NER 2018. For each task, we define the task type (classification, regression, or sequence tagging), the metric used for evaluation, the previous state-of-the-art, and our results including the absolute difference to the SOTA. The competition between XLM-R and our large model dominates the results, since both models have led to significant improvements in linguistic tasks for Polish language. In some cases, the improvement over previous approaches is greater than 10\%. For example, the CDB task was a part of the PolEval 2019 competition, in which the winning solution by \citet{czapla2019universal} achieved an F1-score of 58.6. Both our model and the XLM-R large model outperform that by at least twelve points, achieving an F1-score of over 70. The comparison for the named entity recognition task is also interesting. The previous state-of-the-art solution by \citet{dadas2019combining} is a model that combined neural architecture with external knowledge sources, such as entity lexicons or a specialized entity linking module based on data from Wikipedia. Our language model managed to outperform this method by 3.8 points without using any structured external knowledge. In summary, our model has demonstrated an improvement over existing methods in eleven of the thirteen tasks. 

\section{Conclusions}
We have presented two transformer-based language models for Polish, pre-trained using a combination of publicly available text corpora and a large collection of methodically pre-processed web data. We have shown the effectiveness of our models by comparing them with other transformer-based approaches and recent state-of-the-art approaches. We conducted a comprehensive evaluation on a wide set of Polish linguistic tasks, including binary and multi-class classification, regression, and sequence labeling. In our experiments, the larger model performed better than other methods in eleven of the thirteen cases. To accelerate research on NLP for Polish language, we have released the pre-trained models publicly.

\bibliography{references}

\clearpage
\appendix
\clearpage
\section{Non-English transformer-based language models}

\begin{table*}[!htbp]
  \centering
  \begin{tabularx}{400pt}{llcll}
    \toprule
    \textbf{Project} & \textbf{Languages} & \textbf{Paper} & \textbf{Project URL} \\
    \midrule
Arabic-BERT & Arabic & - & \href{https://github.com/alisafaya/Arabic-BERT}{github.com/alisafaya/Arabic-BERT} \\
AraBERT & Arabic & \citep{antoun2020arabert} & \href{https://github.com/aub-mind/arabert}{github.com/aub-mind/arabert} \\
ClueCorpus2020 & Chinese & \citep{xu2020cluecorpus2020} & \href{https://github.com/CLUEbenchmark/CLUECorpus2020}{github.com/CLUEbenchmark/CLUECorpus2020} \\
Chinese BERT & Chinese & \citep{chinese-bert-wwm} & \href{https://github.com/ymcui/Chinese-BERT-wwm}{github.com/ymcui/Chinese-BERT-wwm} \\
Google BERT & Chinese & \citep{devlin-etal-2019-bert} & \href{https://github.com/google-research/bert}{github.com/google-research/bert} \\
ERNIE 2.0 & Chinese & \citep{sun2019ernie} & \href{https://github.com/PaddlePaddle/ERNIE}{github.com/PaddlePaddle/ERNIE} \\
BERTje & Dutch & \citep{de2019bertje} & \href{https://github.com/wietsedv/bertje}{github.com/wietsedv/bertje} \\
RoBBERT & Dutch & \citep{delobelle2020robbert} & \href{https://ipieter.github.io/blog/robbert}{ipieter.github.io/blog/robbert} \\
Finnish BERT & Finnish & \citep{virtanen2019multilingual} & \href{https://github.com/TurkuNLP/FinBERT}{github.com/TurkuNLP/FinBERT} \\
CamemBERT & French & \citep{martin2019camembert} & \href{https://camembert-model.fr}{camembert-model.fr} \\
FlauBERT & French & \citep{le2019flaubert} & \href{https://github.com/getalp/Flaubert}{github.com/getalp/Flaubert} \\
German BERT & German & - & \href{https://deepset.ai/german-bert}{deepset.ai/german-bert} \\
GreekBERT & Greek & - & \href{https://github.com/nlpaueb/greek-bert}{github.com/nlpaueb/greek-bert} \\
UmBERTo & Italian & - & \href{https://github.com/musixmatchresearch/umberto}{github.com/musixmatchresearch/umberto} \\
GilBERTo & Italian & - & \href{https://github.com/idb-ita/GilBERTo}{github.com/idb-ita/GilBERTo} \\
Japanese BERT & Japanese & - & \href{https://github.com/yoheikikuta/bert-japanese}{github.com/yoheikikuta/bert-japanese} \\
Japanese BERT & Japanese & - & \href{https://github.com/cl-tohoku/bert-japanese}{github.com/cl-tohoku/bert-japanese} \\
KoBERT & Korean & - & \href{https://github.com/SKTBrain/KoBERT}{github.com/SKTBrain/KoBERT} \\
Malaya & Malaysian & - & \href{https://github.com/huseinzol05/Malaya/}{github.com/huseinzol05/Malaya/} \\
Nordic BERT & Nordic (4) & - & \href{https://github.com/botxo/nordic\_bert}{github.com/botxo/nordic\_bert} \\
PolBERT & Polish & - & \href{https://github.com/kldarek/polbert}{github.com/kldarek/polbert} \\
HerBERT & Polish & \citep{rybak2020klej} & \href{https://klejbenchmark.com}{klejbenchmark.com} \\
Portuguese BERT & Portuguese & \citep{souza2019portuguese} & \href{https://github.com/neuralmind-ai/portuguese-bert}{github.com/neuralmind-ai/portuguese-bert} \\
RuBERT & Russian & \citep{kuratov2019adaptation} & \href{https://github.com/deepmipt/DeepPavlov}{github.com/deepmipt/DeepPavlov} \\
SlavicBERT & Slavic (4) & \citep{arkhipov-etal-2019-tuning} & \href{https://github.com/deepmipt/Slavic-BERT-NER}{github.com/deepmipt/Slavic-BERT-NER} \\
BETO & Spanish & \citep{CaneteCFP2020} & \href{https://github.com/dccuchile/beto}{github.com/dccuchile/beto} \\
Swedish BERT & Swedish & - & \href{https://github.com/Kungbib/swedish-bert-models}{github.com/Kungbib/swedish-bert-models} \\
BERTsson & Swedish & - & \href{https://huggingface.co/jannesg/bertsson}{huggingface.co/jannesg/bertsson} \\
BERTurk & Turkish & - & \href{https://github.com/stefan-it/turkish-bert}{github.com/stefan-it/turkish-bert} \\
PhoBERT & Vietnamese & \citep{nguyen2020phobert} & \href{https://github.com/VinAIResearch/PhoBERT}{github.com/VinAIResearch/PhoBERT} \\
    \bottomrule
  \end{tabularx}
  \label{tab:bert_models}
\end{table*}

\end{document}